\documentclass{INTERSPEECH2023}

\usepackage{CJKutf8}

\usepackage{times}
\usepackage{latexsym}

\usepackage{microtype}
\usepackage{hyperref}
\usepackage{tabularx}
\usepackage{graphicx}
\usepackage{xspace}
\usepackage{paralist}
\usepackage{booktabs}
\usepackage{multirow}
\usepackage{amsmath}
\usepackage{amsfonts}
\usepackage{amssymb}
\usepackage{pifont}
\usepackage{mathrsfs}
\usepackage{algorithm,algpseudocode}
\usepackage{mdwlist}
\usepackage{enumitem}
\usepackage{color}
\usepackage{dsfont}
\usepackage{amsthm}
\usepackage{bm}
\usepackage{subfig}
\usepackage[dvipsnames]{xcolor}
\usepackage{soul}
\usepackage{makecell}
\usepackage{bbding}

\newcommand{\notes}[1]{}

\theoremstyle{definition}

\theoremstyle{plain}

\newcommand{\ith}[1]{\ensuremath{i^{{th}}}}

\newcount\permx
\newcount\permy
\def\permdot#1#2{
\permx=#1 \advance\permx by-1
\permy=#2 \advance\permy by-1
\psframe[fillcolor=black, fillstyle=solid]
(\permx,\permy)(#1, #2)
}

\newcommand{\boxnum}[1]{{\setlength{\fboxsep}{1pt}\raisebox{1pt}{\hspace{1pt}\fbox{\tiny #1}\hspace{1pt}}}}
\newcommand{\ind}[1]{\ensuremath{_{\kern-0.5pt\boxnum{#1}}}}

\newcommand{\smallnt}[1]{\ensuremath{_{\mbox{\tiny PP}}}\xspace}

\newcommand{\pseudocode}{Algorithm}
\floatname{algorithm}{\pseudocode}

\iffalse

\else

\fi

\newcommand{\newdataset}{GigaST\xspace}

\interspeechcameraready 

\title{GigaST: A 10,000-hour Pseudo Speech Translation Corpus}
\name{
    Rong Ye*,
    Chengqi Zhao*
        \thanks{* Equal contribution.},
    Tom Ko,
    Chutong Meng\textsuperscript{\textdagger}
        \thanks{\textsuperscript{\textdagger} Work is done during internship at Bytedance.},
    Tao Wang,
    Mingxuan Wang,
    Jun Cao
}

\address{
  ByteDance
}
\email{
\texttt{\{yerong, zhaochengqi.d, tom.ko, wangtao.960826, wangmingxuan.89, caojun.sh\}@bytedance.com, mengct00@gmail.com}
\texttt{}
}

\begin{document}

\maketitle
 
\begin{abstract}

This paper introduces GigaST, a large-scale pseudo speech-to-text translation (ST) corpus.
We create the corpus by translating the transcript in GigaSpeech, an English ASR corpus, into German and Chinese. 
The training set is translated by a strong machine translation system and the test set is translated by human. 
ST models trained with an addition of our corpus obtain new state-of-the-art results on the MuST-C English-German benchmark test set.
We provide a detailed description of the translation process and verify its quality.
We make the translated text data public and hope to facilitate research in speech translation. 
Additionally, we also release the training scripts on NeurST\footnote{\url{https://github.com/bytedance/neurst/}} to make it easy to replicate our systems.
\newdataset dataset is available at \url{https://st-benchmark.github.io/resources/GigaST}.
\end{abstract}

\section{Introduction}

End-to-end speech-to-text translation (ST) directly translates the speech in the source language into sentences in the target language, without outputting the source language transcription~\cite{berard2016listen}.
With the success of attention-based models for speech and text-related tasks, a typical and effective baseline model for ST is speech-transformer~\cite{dong2018speech,inaguma2020espnet,wang2020fairseqs2t, zhao2021neurst}, which has much of the same model structure as the commonly used MT model Transformer~\cite{duong2016attentional}, except for the pre-processing down-sampling module for speech signals.

To train such an end-to-end ST model, a high-quality dataset is important.
In general, the more data available, the better the model can be trained. For example, in the MT task, as the bilingual parallel data increases, so does the translation performance.
As for the speech translation benchmark dataset MuST-C English-German (En-De), which is currently most widely compared by various models, for example, there are only 234k samples (408 hours), while there are more than 4M parallel text training samples for En-De text translation, and in comparison, they are not even in the same order of magnitude.
If the ST training data is also upgraded to the same order of magnitude as the MT data, what will be the translation performance of various end-to-end models? 


Therefore, in this paper, we try to build a large-scale speech-to-text translation dataset. Fortunately, GigaSpeech~\cite{chen2020giga} provided 10,000 hours of English \textit{speech-transcription} parallel data, which served as the speech source for our datasets.
We extend the GigaSpeech ASR dataset to a massive ST corpus -- GigaST, up to 25 times as large as the existing open-source dataset, such as MuST-C~\cite{digangi2019must} and TEDx~\cite{salesky2021multilingual}.
Specifically, the target-side translations of the training set are obtained by translating the transcription using high-quality MT models.
The translations in the test sets are manually annotated and verified one by one, which avoided the alignment errors in the previous MuST-C dataset.
The GigaST dataset contains both English-Chinese (En-Zh) and English-German (En-De) translation directions.

Using this dataset, we first evaluate the performance of the standard Speech-Transformer models.
Then we evaluate the performance of SSL-Transformer models (SSL stands for self-supervised learning) by incorporating pre-trained speech encoders, Wav2vec2~\cite{baevski2020wav2vec} and HuBERT~\cite{hsu2021hubert}.
Results show that increasing the size of the training dataset with GigaST improves the BLEU scores in all test sets.





\section{Dataset Creation}

This section describes our method of creating a pseudo ST corpus from an existing ASR corpus. Pseudo-labeling has been proven effective in various machine learning tasks~\cite{akhbardeh2021wmt, yejia2019pseudo, KULIGOWSKA20211162}.

\subsection{Training Set}
\label{sec:train_set}
We start from GigaSpeech~\cite{chen2020giga}, a multi-domain English speech recognition corpus with 10,000 hours of labeled audio, and create paired text-to-text translation with an MT model.
To obtain high-quality pseudo labels, we train deep transformer-based machine translation models~\cite{vaswani2017attention}, with 24 layers of the encoder and 6 layers of the decoder. 
The training data for MT consists of  WMT2021~\footnote{\url{https://www.statmt.org/wmt21/translation-task.html}} and CCMatrix, CCAlign and OpenSubtitles portions from OPUS~\footnote{\url{https://opus.nlpl.eu/}}. 
We follow the data filtering and pre-processing methods described in \cite{li2019niutrans,wu2020volctrans}, and utilize iterative sequence-level knowledge distillation~\cite{kim2016kd,freitag2017kd} and back translation~\cite{sennrich2016bt} techniques to improve the performance of MT. 

The utterances of the original GigaSpeech corpus are segmented from long audio recordings.
The discourse phenomena, such as pronominal anaphora and lexical consistency will be neglected by sentence-level MT systems, which makes it far worse than human translations~\cite{bawden2018evaluating,laubli2018doceval}.
Therefore we apply \textit{multi-resolution training}~\cite{sun2020doctrans} to build a context-aware MT system. 
Specifically, we first concatenate each segment with its previous one and insert a special token \texttt{[SEP]} as the segment separator. 
Then the generated translation is split according to the \texttt{[SEP]} token.

Since our training dataset is artifacted from GigaSpeech ASR corpus based on the MT system described above, we want to verify how good the MT system is and whether it is close to the real translations of translators.
To this end, we perform a comparison in terms of both automated metrics evaluation and human evaluation.

Table~\ref{tab:teacher_mt_bleu} illustrates the automated metrics evaluation (in BLEU) of our MT models for En-Zh and En-De directions on \textit{newstest2021} set, where Online-B/W and Online-W/A are top-2 online systems individually for En-Zh and En-De reported by \cite{akhbardeh2021wmtfindings}. 
It shows that our MT model can obtain state-of-the-art performance and is comparable with online systems.

\begin{table}[ht]
    \caption{BLEU scores of our MT models versus online MT systems on \textbf{newstest2021} test set.}
    \setlength{\abovecaptionskip}{-0.2cm}
    \centering
    \begin{tabular}{c|c|c}
        \toprule
        \textbf{Direction} & \textbf{System} & \textbf{BLEU} \\
        \midrule
        \multirow{3}{*}{En-Zh}& Online-B & 48.5 \\
        & Online-W & 44.8 \\
        & Our model & 47.3 \\
        \midrule
        \multirow{3}{*}{En-De}& Online-W & 51.0 \\
        & Online-A & 47.6 \\
        & Our model & 49.8 \\
        \bottomrule
    \end{tabular}
    \label{tab:teacher_mt_bleu}
\end{table}

Furthermore, we conduct the human evaluation on the pseudo labels to verify the quality. 
First, we sample 30 audios from the training set and randomly select 1 to 20 continuous segments from each audio, with a total of 320 unique segments.
Then 2 professional translators separately produce Chinese and German translations for this evaluation set, which we take as the ground truth. And another 6 evaluators (3 for En-Zh and 3 for En-De) are asked to rate the translations of both human and MT systems from 0 to 6. A rating of 6 indicates that the expression is fluent and the meaning of the translation is faithful to the source without any grammar errors, while a rating of 0 means the translation is incomprehensible and full of errors. Table~\ref{tab:data_human_eval} 
lists the averaged scores from the evaluators. The En-Zh MT system gets a rating of 4.14 and for En-De it is 4.82. The above 4 rating means our generated translations are semantically consistent with the source texts. The weakness mainly comes from fluency problems, such as unidiomatic word translations, informal expressions, and function word errors. Moreover, the rating of En-De MT (4.82) is close to that of human translations (5.06), which further verifies the quality of our produced training set.

\begin{table}[ht]
    \caption{Human evaluation results on the translations by human translators and machine translation models.}
    \setlength{\abovecaptionskip}{-0.3cm}
    \centering
    \begin{tabular}{c|cc}
        \toprule
        \textbf{Direction} & \textbf{Human} & \textbf{MT}  \\
        \midrule
        En-Zh & 4.92 & 4.14 \\
        En-De & 5.06 & 4.82 \\
        \bottomrule
    \end{tabular}
    \label{tab:data_human_eval}
\end{table}

\subsection{Test Sets}
\label{sec:test_set}
We provide En-Zh and En-De test sets in \newdataset. 
The En-Zh test set contains the translation of all GigaSpeech test utterances while the En-De test set contains a subset of it for this current release. 
The test sets are produced by human translators looking at the transcriptions. 
A small number of transcriptions are difficult to understand due to the lack of context and are ignored in our test sets.

The statistics of \newdataset are listed in Table~\ref{tab:data_stats}. Note that non-speech segments, such as music and noise, are not included in our statistics. 
The \#tokens is counted in character and word for En-Zh and En-De respectively.
\begin{table}[t]
    \caption{The statistics of \newdataset}
    \setlength{\abovecaptionskip}{-0.5cm}
    \centering
    \begin{tabular}{c|c|r|r|r}
        \toprule
        \textbf{Direction} & \textbf{Subset} & \multicolumn{1}{c|}{\textbf{\#seg.}} & \multicolumn{1}{c|}{\textbf{\#hours}} & \multicolumn{1}{c}{\textbf{\#tokens}} \\
        \midrule
        \multirow{5}{*}{En-Zh}
        & S & 210,012 & 243.1 & 4.1M \\
        & M & 835,846 & 974.3 & 16.7M  \\
        & L & 2,084,274 & 2,337.7 & 41.8M \\
        & XL & 7,650,889 & 9,780.8 & 168.3M \\
        & Test & 19,888 & 35.3 & 0.6M \\
        \midrule
        \multirow{5}{*}{En-De}
        & S & 221,572 & 256.2 & 2.5M \\
        & M & 868,316 & 1,013.1 & 10.2M \\
        & L & 2,147,471 & 2,510.9 & 25.3M \\
        & XL & 7,815,436 & 9,997.9 & 101.6M\\
        & Test & 4,163 & 7.1 & 0.7M\\
        \bottomrule
    \end{tabular}
    
    \label{tab:data_stats}
\end{table}

\section{Experiment}
    
    
    
    

\begin{table*}[t]
    \caption{\textbf{En-Zh BLEU scores.} 
    \textbf{S, M, L} and \textbf{XL} stand for training data of various scales.
    The test set includes \newdataset test set as created in Section~\ref{sec:test_set} and the in-house test set. }
    \small
    \centering
    \begin{tabular}{cl|c|cccc|cccc}
        \toprule
        
        & & &  \multicolumn{4}{c|}{\textbf{\newdataset Test}} &
        \multicolumn{4}{c}{\textbf{In-house Test}} \\
        \cmidrule{4-11}
        \multicolumn{2}{c|}{\textbf{Models}}& 
        \textbf{\# params} &
        \textbf{S} & \textbf{M} & \textbf{L} & \textbf{XL} & \textbf{S} & \textbf{M} & \textbf{L} & \textbf{XL}\\
        
        \midrule\midrule
        \multirow{3}{*}{\makecell[c]{Speech-\\Transformer}}
        & \texttt{S-Transf\_S} & 37 M &
            24.2 & 28.8 & 29.8 & 29.9 & 20.1 & 24.6 & 25.5 & 25.9 \\
        & \texttt{S-Transf\_M} & 90 M &
            23.3 & 31.4 & 33 & 33.6 & 19.4 & 26.6 & 28.5 & 29.4  \\
        & \texttt{S-Transf\_L} & 322 M &
            21.2 & 31.4 & 35 & 36.3 & 17.8 & 26.7 & 30.2 & 31.5 \\
        \midrule
        
        & \texttt{w2v2-base} & 359 M & 
            27.2 & 32.1 & 34.2 & 37.4 &
            22.1 & 27.0 & 28.8 & 32.1 \\
        SSL- & \texttt{w2v2-large} & 581 M & 
            27.0 & 31.6 & 33.9 & 36.9 & 
            19.8 & 24.5 & 26.9 & 30.4 \\
        Transformer & \texttt{hubert-base} & 359 M &
            27.6 & 31.8 & 34.0 & 37.2 & 
            22.5 & 25.9 & 29.3  & 32.1 \\
        & \texttt{hubert-large} & 581 M &
            \textbf{30.1} & \textbf{33.4} & \textbf{35.6} & \textbf{38.0} & 
            \textbf{24.4} & \textbf{27.9} & \textbf{30.3} & \textbf{32.5} \\
        \bottomrule
    \end{tabular}
    \label{tab:ST_zh}
\end{table*}

    
    
    

\begin{table*}[t]
    \setlength{\belowcaptionskip}{-0.2cm}
    \caption{\textbf{En-De BLEU scores.} The training sets include \textbf{S/M/L/XL} subsets plus MuST-C En-De training set. The test sets are \newdataset En-De test set and MuST-C \texttt{tst-COMMON} set.}
    \small
    \centering
        
        
    \begin{tabular}{cl|c|cccc|cccc}
        \toprule
        & & &  \multicolumn{4}{c|}{\textbf{\newdataset Test}} &
        \multicolumn{4}{c}{\textbf{MuST-C tst-COM}} \\
        \cmidrule{4-11}
        \multicolumn{2}{c|}{\textbf{Models}}& 
        \textbf{\# params} &
        \textbf{S} & \textbf{M} & \textbf{L} & \textbf{XL} & \textbf{S} & \textbf{M} & \textbf{L} & \textbf{XL}\\
        
        \midrule\midrule
        \multirow{3}{*}{\makecell[c]{Speech-\\Transformer}}
        & \texttt{S-Transf\_S} & 35 M & 
            21.5 & 24.1 & 25.1 & 25.6 & 24.4 & 25.7 & 25.8 & 25.1 \\
        & \texttt{S-Transf\_M} & 87 M & 
            22.7 & 27.3 & 28.8 & 29.6 & 24.9 & 27.6 & 28.2 & 28.4 \\
        & \texttt{S-Transf\_L} & 315 M & 
            21.3 & 27.5 & 31.1 & 32.6 & 23.2 & 27.8 & 29.6 & 30.1 \\
        \midrule
        
        & \texttt{w2v2-base} & 359 M &
            24.3 & 28.0 & 32.1  & 33.4 &
            26.5 & 28.0 & 29.9 & 30.3\\
        SSL- & \texttt{w2v2-large} & 581 M &
            23.6 & 28.5 & 28.5 & 33.0 &
            23.4 & 26.9 & 27.0 & 30.2 \\
        Transformer & \texttt{hubert-base} & 359 M &
            24.1 & 28.3 & 30.2 & 33.7 &
            24.9 & 27.7 & 29.2 & 30.5 \\
        & \texttt{hubert-large} & 581 M &
            \textbf{27.1} & \textbf{30.6} & \textbf{31.8} & \textbf{33.5} &
            \textbf{27.7} & \textbf{29.6} & \textbf{30.1} & \textbf{30.6} \\
        \bottomrule
    \end{tabular}
    
    \label{tab:ST_de}
\end{table*}
In this section, we conduct ST experiments with speech transformer models and SSL-Transformer models. All models are implemented using NeurST~\cite{zhao2021neurst}.

\subsection{Setups}

\noindent\textbf{Preprocessing and Filtering}~
For speech transformer models, we extract 80-channel log-Mel filterbank coefficients (fbank) of the audio, with windows of 25ms and steps of 10ms, and then apply CMVN (cepstral mean and variance normalization). The SSL-Transformer models use raw wave signals as the input. 
For the text side, words are encoded in subword-level. 
In detail, we lowercase the English transcriptions, remove all punctuations and use SentencePiece\footnote{\url{https://github.com/google/sentencepiece}} with a vocabulary of 15,000. 
For Chinese text, we first segment sentences by Jieba\footnote{\url{https://github.com/fxsjy/jieba}}, and then apply Byte-Pair Encoding (BPE)\footnote{\url{https://github.com/rsennrich/subword-nmt}}~\cite{sennrich2016neural} with 32,000 merge operations.
German texts are first tokenized using Moses tokenizer, followed by BPE with 32,000 merge operations.
During training, we truncate the audio to 30 seconds for GPU memory efficiency, that is, 480,000 for raw wave signals or 3,000 fbank frames. 
We remove training samples whose translation text longer than 120 tokens or the percentage of aligned words to the original English transcription is less than 40\% (produced by \texttt{fast-align}\footnote{\url{https://github.com/clab/fast_align}} toolkit). 


\noindent\textbf{Training} For En-Zh, we use \newdataset as the training set and use TED \texttt{dev2010} and \texttt{tst2015}~\cite{liu2019end} as the validation set. For En-De, MuST-C is added to the training set and we use MuST-C \texttt{dev} set for validation.



\noindent\textbf{Evaluation}~
Apart from the GigaST test set described in Section~\ref{sec:test_set}\ref{sec:test_set}, for both language directions, we add two additional test sets: an in-house test set containing a total of 8.5 hours of news and tech talks (3,917 sentences) for En-Zh and MuST-C \texttt{tst-COMMON} set for En-De.
The metric we use is case-sensitive detokenized BLEU\footnote{\url{https://github.com/mjpost/sacrebleu}}.

\subsection{End-to-end Models}
\label{sec:e2e_setups}
We compare various end-to-end ST models of different sizes.

\noindent\textbf{Speech-Transformer}~\cite{dong2018speech} is our benchmark model. 
The model uses fbank features as the input, and stacks two $3 \times 3$ CNN layers with stride size 2 and a transformer encoder-decoder module.
Specifically, we implement three different model sizes, namely \texttt{S-Transf\_S}, \texttt{S-Transf\_M}, \texttt{S-Transf\_L}, with model hyper-parameters listed in Table~\ref{tab:strans_hyperparam}.
We follow the setup of the optimizer and the learning rate schedule as in \cite{zhao2021neurst}, as well as using ASR pre-training and SpecAugment~\cite{park2019specaugment}.

\begin{table}[ht]
    \setlength{\belowcaptionskip}{-0.2cm}
    \caption{Hyper-parameters for Speech-Transformer  models}
    \centering
    \begin{tabular}{l|ccc}
        \toprule
        & {\texttt{S}} & {\texttt{M}} & {\texttt{L}} \\
        \midrule
        Hidden Size & 256 & 512 & 1024 \\
        Filter Size & 2048 & 2048 & 4096 \\
        Attention Heads & 4 & 8 & 16 \\
        Encoder Layers & 12 & 12 & 12\\
        Decoder Layers & 6 & 6 & 6 \\
        \bottomrule
    \end{tabular}
    \label{tab:strans_hyperparam}
\end{table}

\noindent\textbf{SSL-Transformer}~ 
As recent research on self-supervised learning (SSL) in speech has intensified, pre-trained speech encoders, such as Wav2vec2~\cite{baevski2020wav2vec} and HuBERT~\cite{hsu2021hubert}, have been applied to downstream tasks instead of spectral features~\cite{yang2021superb}. We can adopt a similar idea for the ST task. 
We evaluate SSL-Transformer, which replace Fbank features in Speech-Transformer with representations extracted from pre-trained speech encoders. To reduce the sequence length, we add two layers of convolutional subsampler with stride=2 after the SSL module, and apply the Transformer encoder-decoder as the downstream module. 
For the SSL speech encoders, we try four of them which performed well on the SUPERB leaderboard\footnote{\url{https://superbbenchmark.org/leaderboard}}, namely \texttt{w2v2-base}, \texttt{w2v2-large}, \texttt{hubert-base}, and \texttt{hubert-large}. 
For downstream Transformer, we use the same hyperparameter as \texttt{S-Trans\_L}, with 6 layers of encoder and decoder, $d_{\text{model}}=1024, d_{\text{ff}}=4096, d_{\text{head}}=8$.
Combining the above modules, we get four models with model parameter sizes of 359M, 581M, 359M, and 581M, respectively.
For the subsequent model notations, since the structures of CNN and Transformer modules are the same, for simplicity, we only use the name of the speech encoders as the name of the whole model.
The raw waveform of the entire speech is fed into the model, and SSL modules are \textbf{NOT} frozen during training. We set warmup steps at $25,000$ and peak learning rate at $2e^{-4}$.

\noindent\textbf{Results}~
The BLEU scores of the En-Zh and En-De test sets are shown in Tables~\ref{tab:ST_zh} and~\ref{tab:ST_de} for different models trained with varying training sets.
It is obvious that, for every models, the performance improves as the training data size increases.
And model capacity often determines how good the translation is.
In general, with the same amount of training data, the larger the model size, the better the performance. 
It is interesting to note that \texttt{S-Transf\_S}, with only 35M parameters, fails to improve substantially as the data size increases from 1k to 10k hours.
On the other hand, large models tend to underfit when the training set is small. For example, when training with the GigaST\_S subset, \texttt{S-transf\_L} performs worse than \texttt{S-transf\_S}.

\subsection{Cascade Systems}
By concatenating the ASR models and the MT models, we obtain cascade systems.
Specifically, our ASR model has the same structure as \texttt{S-transf\_L} and we get 11.8 WER on the GigaSpeech test set, which is on par with other ASR systems on the GigaSpeech leaderboard\footnote{\url{https://github.com/SpeechColab/GigaSpeech}}.
For the MT model, we provide two models, one is Transformer-large trained using GigaSpeech transcription-translation bilingual text, noted as \textit{constrained}, and the other is the MT model introduced in Section~\ref{sec:train_set}, noted as \textit{unconstrained}.
The former is for a fair comparison between the cascade systems and the end-to-end models with the same training data, while the latter has a stronger translation performance with BLEU scores higher than 40.

Table~\ref{tab:cascade} shows the performances of the cascade systems. When the models are trained with the same amount of data, the performance of different end-to-end models are better than those of the cascade systems.
However, the unconstrained cascade models, boosted by the larger amount of text training data, have a higher quality of text translation, which leads to better speech translation than the end-to-end models. 
Now with a powerful MT model,
the gap in BLEU between end-to-end and cascade models is around 2. We still have room to improve the end-to-end performance through pre-training of decoder, multi-task training and other techniques.
We leave these for future work.


\begin{table}[ht]
    \caption{The BLEU scores of cascade models on GigaST test sets}
    \centering
    
    \begin{tabular}{c|lc|c}
        \toprule
        & \multicolumn{2}{c|}{\textbf{MT}} & \textbf{ST} \\
        \textbf{Direction} & \textbf{Condition} & \textbf{BLEU} & \textbf{BLEU} \\
        \midrule
        \multirow{2}{*}{En-Zh} & constrained & 24.9 & 22.3 \\
        & unconstrained & 44.3 & 39.8 \\
        \midrule
        \multirow{2}{*}{En-De} & constrained & 37.6 & 33.4 \\
        & unconstrained & 42.2 & 35.7 \\
        \bottomrule
    \end{tabular}
    
    \label{tab:cascade}
\end{table}

\subsection{Analysis on Speech Representations}

In addition to the results of the baseline Speech-Transformers, we also investigate the performance of pre-trained speech encoders under large-scale speech translation. 
Are these pre-trained encoders still effective and helpful under large amounts of training data? 

Before answering this question, we need to figure out: 
how should these pre-trained speech encoders be incorporated into the training?
Should they be frozen with parameters not updated, or should the whole model be fine-tuned based on ST supervised data?
Taking two pre-trained encoders, \texttt{hubert-base} and \texttt{hubert-large}, as examples, Table~\ref{tab:freeze} shows the results of En-Zh translation with and without the encoder frozen.
It shows that with both base and large models, the frozen case perform much worse than the unfrozen case, with an average difference of as much as 2.6 BLEU.
This differs from the practice of freezing speech encoders in the downstream ASR task, where freezing these encoders can still yield a word error rate of as low as 3.4 on LibriSpeech~\cite{yang2021superb}, but in ST tasks, we recommend fine-tuning the speech encoders together with other components, which can greatly improve ST performance. Therefore, we conduct the rest of our experiments without freezing the speech encoders.
Detailed setups are introduced in Section~\ref{sec:e2e_setups}.

\begin{table}[ht]
    \caption{To freeze or not to freeze the SSL speech encoders? \texttt{hubert-base} and \texttt{hubert-large} were finetuned with \newdataset XL subset and tested on the En-Zh test sets.}
    \centering
    \begin{tabular}{l|c|cc|c}
        \toprule
        \textbf{SSL repr.} & \textbf{freeze?} & \textbf{GigaST} & \textbf{In-house} & \textbf{Avg.} \\
        \midrule
        \multirow{2}{*}{\texttt{hubert-base}} &
        \Checkmark & 34.4 & 29.7 & 32.1 \\
        & \XSolidBrush & 37.2 & 32.1 & 34.7 \\
        \midrule
        \multirow{2}{*}{\texttt{hubert-large}}
        & \Checkmark & 35.3 & 30.0 & 32.7 \\
        & \XSolidBrush & \textbf{38.0} & \textbf{32.5} & \textbf{35.3} \\
        \bottomrule
    \end{tabular}
    
    \label{tab:freeze}
\end{table}


The results of En-Zh and En-De experiments are summarized in Tables~\ref{tab:ST_zh} and \ref{tab:ST_de}.
It can be seen that SSL-Transformer models are generally better than the Speech-Transformer models.
Even though \texttt{S-Transf\_L} and \texttt{w2v2-base} have roughly the same order of magnitude of parameters (300m parameters), \texttt{w2v2-base} outperforms \texttt{S-Transf\_L}.
When the training data size is increased to 10,000 hours, according to the BLEU scores throughout the XL columns, the pre-trained speech module continues to play an essential role in improving ST performance.
In addition, looking at the performances of the models trained with the S subset (only 250 hours), pre-trained speech encoders are particularly useful.
Take En-Zh translation as an example (Table~\ref{tab:ST_zh}), using \texttt{hubert-Large} to train 250 hours of speech, the translation performance can reach or even exceed \texttt{S-Transf\_S} training on 10,000 hours.
Meanwhile, as the data size increases, we find that the gain from the pre-trained speech encoders for the downstream ST task becomes smaller.
Despite the reduced benefit, 
we observe that pre-trained speech encoders are still useful in our setup.


On the other hand, there are performance differences between the four pre-trained speech encoders. Analyzing the performance of En-Zh and En-De translations comprehensively, the rank among the four is \texttt{hubert-large}, \texttt{hubert-base}, \texttt{w2v2-base}, and \texttt{w2v2-large}. It is surprising to see that \texttt{w2v2-large} (581M parameters) performs worse than  the base models (359M parameters), where it violates the common belief that larger model capacity means more representation capability and with better performance.
Overall, HuBERT models empirically perform better than Wav2vec2 models in our setup.


\section{Conclusion}

This paper presents GigaST, a new speech-to-text corpus suitable for training and evaluating ST systems.
Our corpus is created by translating the transcript in GigaSpeech, which is one of the largest open-source English ASR corpora.
We have demonstrated that models trained with an addition of our corpus can obtain new state-of-the-art results on the MuST-C English-German benchmark test set.
We also establish new benchmark test sets for the two language directions.
We release all the training data, human-translated test sets and our training scripts so that others can easily replicate our results.
We believe our released dataset will open new avenues in speech translation research.

\clearpage
\bibliographystyle{IEEEtran}
\bibliography{ref.clean}

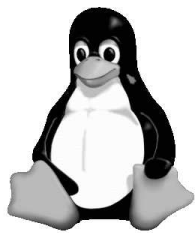
\end{document}